\pdfoutput=1
\documentclass[conference]{IEEEconf}
\IEEEoverridecommandlockouts
\usepackage{cite}
\usepackage{amsmath,amssymb,amsfonts}
\usepackage{algorithmic}
\usepackage{graphicx}
\usepackage{textcomp}
\usepackage{xcolor}
\usepackage{multicol}
\usepackage[bookmarks=true]{hyperref}
\newcommand{\loosepar}{\looseness=-1}
\begin{document}

\title{Learning Contact-based Navigation in Crowds}

\author{Kyle Morgenstein$^{1,C}$
\and
Junfeng Jiao$^2$
\and
Luis Sentis$^1$
\thanks{$^{1}$Department of Aerospace Engineering and Engineering Mechanics, University of Texas at Austin. $^{2}$School of Architecture, University of Texas at Austin. $^{C}$Corresponding Author: kylem@utexas.edu. }
}

\maketitle

 Navigation strategies that intentionally incorporate contact with humans (i.e. “contact-based” social navigation) in crowded environments are largely unexplored even though collision-free social navigation is a well studied problem. Traditional social navigation frameworks require the robot to stop suddenly or “freeze” whenever a collision is imminent \cite{sathyamoorthy2020frozone}. This paradigm poses two problems: 1) freezing while navigating a crowd may cause people to trip and fall over the robot, resulting in more harm than the collision itself, and 2) in very dense social environments where collisions are unavoidable, such a control scheme would render the robot unable to move and preclude the opportunity to deploy robots in common human environments such as public transportation systems. However, if robots are to be meaningfully included in crowded social spaces, such as busy streets, subways, stores, or other densely populated locales, there may not exist trajectories that can guarantee zero collisions. Thus, adoption of robots in these environments requires the development of minimally disruptive navigation plans that can safely plan for and respond to contacts. We propose a learning-based motion planner and control scheme to navigate dense social environments using safe contacts for an omnidirectional mobile robot. Detailed information and additional materials can be found on our project page: \href{https://sites.google.com/view/hirl-contact-based-navigation/}{https://sites.google.com/view/hirl-contact-based-navigation/} \loosepar{}

\begin{figure*}[h!]
\centering
\includegraphics[width=0.8\textwidth]{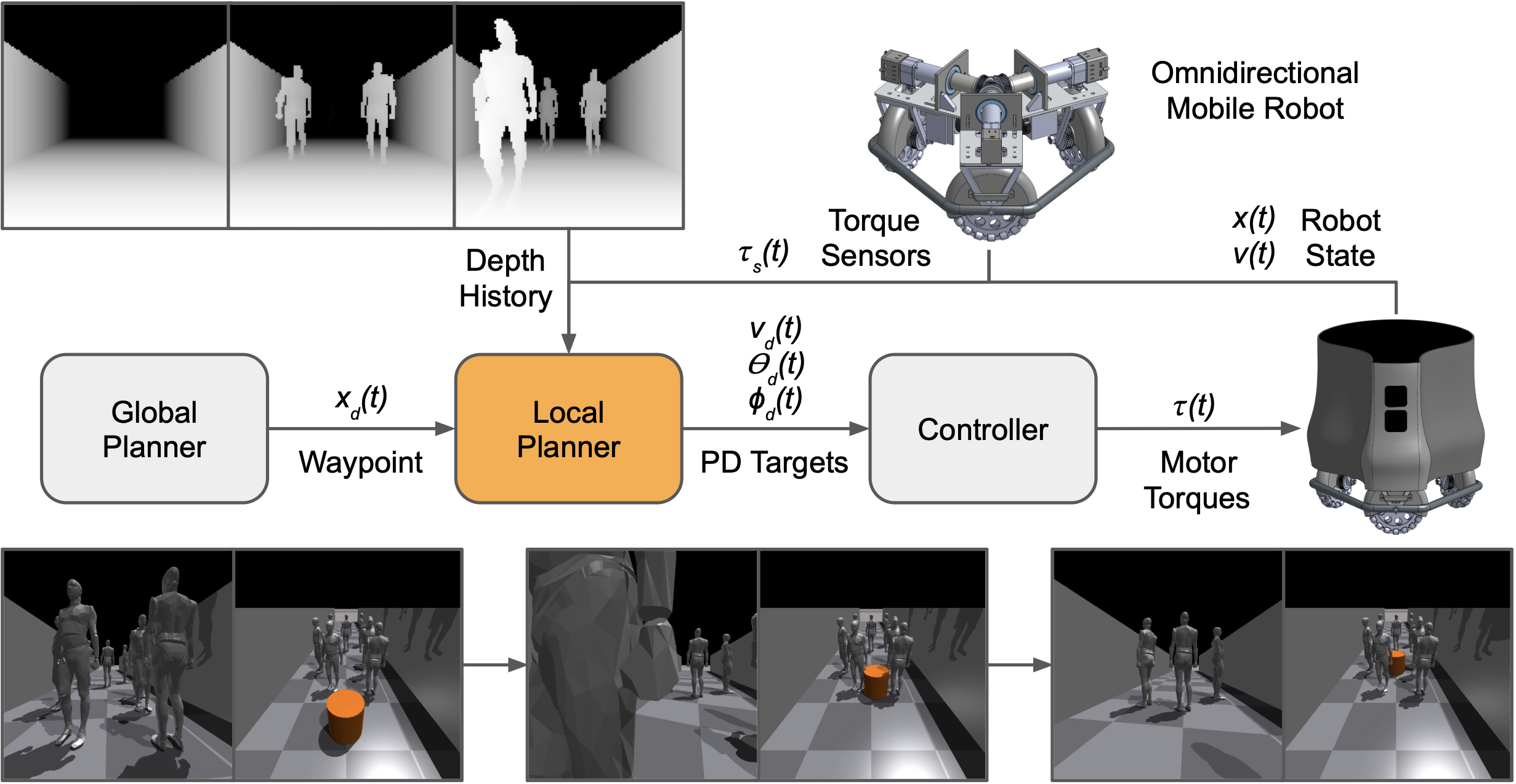}
\caption{Contact-based social navigation. The local planner ingests waypoints from a global planner, robot state information, contact forces, and a history of depth images. The local planner then outputs desired speed, motion heading, and camera heading, which produces the desired velocity in world coordinates as $\dot{\textbf{x}}_d(t) = v_d\big(\begin{smallmatrix}
  \cos{\theta_d}\\
  \sin{\theta_d}
\end{smallmatrix}\big)$. A PD controller then stabilizes the robot’s velocity and heading around the desired setpoint.}
\label{fig:pipeline}
\end{figure*}

Contact-based social navigation remains broadly understudied. One recent attempt to overcome the freezing-robot problem by allowing for safe collisions was to hand-design a controller that allowed the robot to slide around the point of collision into free space \cite{paez2022unfreezing}. This resulted in an effective reactionary controller with maximum contact force guarantees much below human injury-causing thresholds \cite{iso2016ts}. However, this control scheme permits only a single point of contact with the robot and free space for the robot to slide into - both assumptions that may not be satisfied in real social environments. In follow-up work incorporating the control scheme into a larger navigation framework, the highest density tested was $<$ 1 person per square meter (pp$/m^2$) \cite{paez2022pedestrian}. Another study \cite{Shrestha2015UsingCI} investigates the use of intentional touches by a robot arm mounted on a mobile base to encourage people to step out of the robot’s path. However, this study was not designed for crowded environments where no free path can be planned. While these works offer initial solutions to contact-based navigation, the rigidity of these approaches are unlikely to be suitable for deployment in subways, entertainment venues, busy corridors, or other locations with high density traffic.\loosepar{}

Data-driven methods have found success at collision avoidance in the social navigation literature \cite{mirsky2021conflict}. While some methods learn from a model-based or mechanistic description of human behavior such as the Social Force Model \cite{helbing1995social} \cite{ferrer2013robot} \cite{mavrogiannis2016interpretation} \cite{fisac2019hierarchical} \cite{fridovich2020confidence}, others take a purely model-free approach to human behavior estimation \cite{tolani2021visual} \cite{bansal2020combining} \cite{seo2022learning}. However, to our knowledge, learning methods have not yet been applied to contact-based navigation in dense social environments. 

To reconcile these limitations, we devise a learning-based motion planner and control scheme that effectively navigates dense social environments. Critically, our planner makes no assumptions about the dynamics of obstacles in the environment, nor does our planner prescribe an “optimal” contact. Instead, the local planner learns an implicit representation of both the environment and contact dynamics, and uses this model to estimate desired headings. Given the difficulty of explicitly modeling inter-crowd interactions, this generality allows our approach to successfully navigate crowds of higher density than has been previously reported.\loosepar{}

The local planner is formulated as the solution to a multi-task reinforcement learning problem. We then train the planner via proximal policy optimization (PPO) \cite{schulman2017proximal}. In an effort to avoid prescribing collision avoidance behaviors and collision resolution strategies a priori, the minimal task formulation is given as 1) follow a coarse path from a global planner, and 2) minimize disturbances to humans. We define disturbances as the ratio between measured contact forces and the average pain threshold for blunt impacts between a human and a robot, as established by the ISO 15066:2016 standard. The average pain threshold for the lower legs is 130 Newtons \cite{iso2016ts}. Expressing contact forces as a ratio allows the planner to implicitly reason about what sorts of contacts should be considered tolerable.\loosepar{}

We model our local planner as a neural network policy. In particular, we use a convolutional neural network to extract relevant features from a sequence of images from the robot’s depth camera. These features are downsampled and fused with an observation of the robot’s state before being passed to the multilayer perceptron backbone of the policy. We design the policy to be relatively shallow so that it can be evaluated in real time on the limited robot hardware. The policy has three outputs: desired speed, desired heading of motion, and desired camera heading. Note that the heading and direction of motion are decoupled due to the omni wheels. We extract three additional auxiliary outputs from the convolutional feature extractor. These outputs are used to estimate the distance to the nearest wall to the left and right of the robot and the distance to the nearest human in the camera’s field of view. While not used for control, these outputs are used to train the policy more efficiently so the feature extractor is pushed to learn to recognize humans and other obstacles in the environment. \loosepar{}

We train the local planner massively in parallel in simulation using NVIDIA IsaacGym \cite{makoviychuk2021isaac}. Environments are procedurally generated to mimic various hallways in our robot testing facility. During training, each environment is populated by simulated humans with deformable joints at a density of 1 pp$/m^2$. Goal locations are selected five meters in front of the robot, and an A* path is generated at runtime assuming that the environment is empty. 200 environments are trained for 5,000,000 total timesteps. The control frequency is 10 Hz while the physics frequency is 100 Hz with 4 substeps, resulting in 400 Hz physics updates. Training on a workstation NVIDIA 3080 12GB GPU takes 12 hours.\loosepar{}

The policy is evaluated in simulation over 360 trials with crowd densities varying between 0.0 and 1.6 pp$/m^2$. See Table \ref{tab:data} for a summary of results. We define success as the percent of trials in which the safe contact threshold (130 N) was not violated. In trials with the highest densities, some crowd configurations had no possible safe paths to the target. In those environments, success is achieved if the robot comes to a stop without violating the safe contact constraint. \loosepar{}

\begin{table}[htbp]
\centering
\caption{}
\label{tab:data}
\begin{tabular}{l|ll}
\textbf{Density} (pp$/m^2$) & \textbf{Success} & \textbf{Time to Completion ($\sigma$)} \\ \hline
    $<$1.0  & 96\% & 11.28 (2.32) \\
    $=$1.0  & 90\% & 13.25 (5.37) \\
    $>$1.0  & 73\% & 16.32 (6.94)                          
\end{tabular}
\end{table}

Failures can be primarily attributed to two causes: sensing and mechanical. Sensing failures occur when obstacles lay outside the depth camera’s receptive field. In particular, the robot is unable to observe the feet and lower legs of humans that are $<$ 0.25 meters from the robot due to the orientation and physical limitations of the depth camera. Sensing failures can be resolved by the inclusion of additional sensors, such as a down-angled camera and a rear-facing camera. Mechanical failures occur when the local planner recognizes a human in the robot’s path, commands the robot to stop or reverse direction, but is not able to avoid contact due to actuation constraints. Resolving mechanical failures includes increasing authority in the motor controllers and reducing the maximum speed that the local planner can command the robot. Despite these failure modes, the local planner is able to use contact to safely navigate in crowds of higher density than has been previously reported, to our knowledge.\loosepar{}

To validate our contact-based navigation scheme, we will deploy the motion planner on our research platform, Bumpybot - a mobile base with torque sensing omni wheels. Torque sensors are used to accurately estimate the location, direction, and magnitude of contact forces applied to the robot \cite{kim2016full}, making Bumpybot ideal for studying navigation in collision-prone environments. We will conduct interactive experiments with small crowds of people in our robot testing facility. Each participant will be given a clicker that records when the clicker is pressed. Participants will be instructed to depress the clicker whenever they feel uncomfortable, whether due to contact with the robot, or the behavior of the robot as it navigates through the crowd. We will compare measured contact forces to human-reported comfort with the robot’s behavior. We will additionally collect survey data about the participant’s perception of the robot’s behavior. We hope that this qualitative analysis will help drive future research into robot sociability and compliance with social norms. Robots are posed to enter society en masse - if we hope to unlock their full potential, it is essential that we earn the public’s trust through safe, reliable human-robot interaction. We hope this study helps lay the groundwork to build that future.\loosepar{}

\bibliographystyle{IEEEtran}
\bibliography{ref}

\end{document}